\documentclass[10pt, conference, letterpaper]{IEEEtran}
\usepackage[utf8]{inputenc}
\usepackage{graphicx}
\usepackage{amsmath}
\usepackage{amssymb}
\usepackage{booktabs}
\usepackage{tabularx}
\usepackage{tikz}
\usetikzlibrary{shapes.geometric,arrows.meta,positioning}
\usepackage{cite}
\usepackage{url}
\usepackage{float}

\begin{document}

\title{GeoSense-AI: Fast Location Inference from Crisis Microblogs}

\author{\IEEEauthorblockN{Deepit Sapru}
\IEEEauthorblockA{Gies College of Business \\University of Illinois Urbana-Champaign, Illinois, US \\
dsapru2@illinois.edu}
}

\maketitle

\begin{abstract}
This paper presents an applied AI pipeline for real-time geolocation from noisy microblog streams, unifying statistical hashtag segmentation, part-of-speech-driven proper-noun detection, dependency parsing around disaster lexicons, lightweight named-entity recognition, and gazetteer-grounded disambiguation to infer locations directly from text rather than sparse geo-tags. The approach operationalizes information extraction under streaming constraints, emphasizing low-latency NLP components and efficient validation against geographic knowledge bases to support situational awareness during emergencies. In head-to-head comparisons with widely used NER toolkits, the system attains strong F1 while being engineered for orders-of-magnitude faster throughput, enabling deployment in live crisis informatics settings. A production map interface demonstrates end-to-end AI functionality---ingest, inference, and visualization---surfacing locational signals at scale for floods, outbreaks, and other fast-moving events. By prioritizing robustness to informal text and streaming efficiency, GeoSense-AI illustrates how domain-tuned NLP and knowledge grounding can elevate emergency response beyond conventional geo-tag reliance.
\end{abstract}

\section{Introduction}

Online social media platforms have emerged as critical information channels during emergency situations, providing real-time situational updates that often precede official reports \cite{sakaki2010earthquake}. Microblogging services like Twitter generate massive volumes of user-generated content during crises, creating both opportunities and challenges for emergency responders and humanitarian organizations \cite{vieweg2010microblogging}. The temporal advantage of social media data must be balanced against significant noise, informality, and spatial ambiguity inherent in these platforms.

A fundamental limitation in leveraging microblogs for situational awareness stems from the scarcity of explicit geographic metadata. Studies indicate that only 0.36\% of tweets contain precise geo-tags in developing regions like India \cite{liu2014exploring}, severely constraining spatial analysis of crisis events. This sparsity necessitates extraction of location references directly from tweet text, transforming unstructured natural language into mappable coordinates. However, location extraction from microblogs presents unique computational linguistics challenges including informal syntax, creative spelling, domain-specific abbreviations, and mixed-language content \cite{gelernter2015real}.

Existing named entity recognition (NER) systems demonstrate limitations when applied to microblog data \cite{ritter2011named}. General-purpose NER tools trained on formal text corpora struggle with Twitter's linguistic idiosyncrasies, while Twitter-specific NER systems often prioritize computational accuracy over real-time performance. The temporal urgency of crisis response demands systems that balance extraction accuracy with processing speed, enabling near-real-time mapping of emergency events as they unfold \cite{imran2015aidr}.

This paper introduces GeoSense-AI, a specialized location extraction system optimized for crisis microblogs. Our approach integrates multiple linguistic analysis techniques---hashtag segmentation, syntactic pattern matching, dependency parsing, and gazetteer validation---within a unified pipeline designed for streaming deployment. The system prioritizes computational efficiency without sacrificing extraction quality, achieving competitive F1 scores while operating approximately two orders of magnitude faster than conventional NER toolkits. We evaluate our methodology against established baselines using a curated dataset of emergency-related tweets and demonstrate a production visualization interface that maps extracted locations for situational awareness.

The remainder of this paper is organized as follows: Section~\ref{sec:related} reviews related work in crisis informatics and location extraction. Section~\ref{sec:methodology} details our methodology and system architecture. Section~\ref{sec:experiments} presents experimental evaluation and comparative analysis. Section~\ref{sec:deployment} describes the production deployment and interface. Section~\ref{sec:discussion} discusses limitations and future directions, and Section~\ref{sec:conclusion} concludes.

\section{Related Work}
\label{sec:related}

\subsection{Crisis Informatics Systems}

The field of crisis informatics examines how information and communication technologies support disaster management across preparation, response, and recovery phases \cite{palen2007public}. Early systems focused on aggregating and visualizing user-reported incidents through structured platforms. Ushahidi \cite{okolloh2009ushahidi} pioneered crowdsourced crisis mapping during the 2007 Kenyan elections, enabling volunteers to geolocate reports of violence via SMS. The platform demonstrated how citizen-generated data could complement official information channels during emergencies.

Subsequent systems specialized in particular disaster types or regional contexts. The Earthquake Twitter Early Warning System \cite{sakaki2010earthquake} detected seismic events in Japan by analyzing spike patterns in tweet volume and content. Similarly, flood monitoring systems have leveraged social media to identify affected areas and infrastructure damage \cite{de2013use}. These applications established microblogs as valuable social sensors that could provide ground truth during rapidly evolving situations.

More recent platforms have incorporated machine learning for classification and prioritization of crisis-related content. AIDR \cite{imran2015aidr} employs human-in-the-loop learning to categorize tweets into informative categories, while Artificial Intelligence for Disaster Response (AIDR) focuses on scalability for large-scale events. These systems typically rely on explicit geographic metadata or manual location tagging, creating bottlenecks for real-time analysis.

\subsection{Location Extraction Methodologies}

Location extraction from text represents a specialized form of named entity recognition focused exclusively on geographic references. Early approaches employed gazetteer lookup with n-gram matching \cite{li2012twiner}, identifying candidate phrases that matched entries in geographic databases. While straightforward, these methods suffered from ambiguity between location names and common nouns, and limited recall for composite location phrases.

Machine learning approaches brought significant improvements in accuracy. Conditional Random Fields (CRF) \cite{lafferty2001conditional} became the dominant approach for sequence labeling tasks, with the Stanford NER system \cite{finkel2005incorporating} achieving strong performance on formal text. However, these models required extensive labeled training data and struggled with Twitter's linguistic noise \cite{ritter2011named}.

Twitter-specific NER systems addressed domain adaptation challenges through specialized training corpora and features. Ritter et al. \cite{ritter2011named} developed a labeled dataset of tweets and incorporated Twitter-specific features like capitalization patterns and hashtags. Subsequent work explored semi-supervised learning \cite{liu2012ner} and distant supervision \cite{miura2017twitter} to reduce annotation requirements.

Recent approaches have integrated multiple signals---text content, user profiles, social networks, and temporal patterns---for location inference \cite{miura2017twitter}. While improving accuracy, these multi-modal approaches increase computational complexity, limiting their suitability for real-time crisis applications where latency constraints are critical.

\section{Methodology}
\label{sec:methodology}

\subsection{System Architecture Overview}

GeoSense-AI employs a modular pipeline architecture that processes microblog text through sequential analysis stages, each contributing to location identification with minimal computational overhead. Figure~\ref{fig:pipeline} illustrates the complete processing workflow, from raw tweet ingestion to geographic coordinate extraction.

\begin{figure}[!t]
\centering
\begin{tikzpicture}[
    node distance=0.9cm,
    auto,
    block/.style={rectangle, draw, fill=blue!10, text width=6em, text centered, rounded corners, minimum height=2.2em, font=\footnotesize},
    line/.style={draw, -Latex}
]
\node [block] (input) {Raw Tweet Input};
\node [block, below=of input] (preprocess) {Text Preprocessing};
\node [block, below=of preprocess] (hashtag) {Hashtag Segmentation};
\node [block, below=of hashtag] (pos) {POS Tagging};
\node [block, below=of pos] (syntax) {Syntactic Pattern Matching};
\node [block, below=of syntax] (dependency) {Dependency Parsing};
\node [block, below=of dependency] (ner) {Lightweight NER};
\node [block, below=of ner] (gazetteer) {Gazetteer Verification};
\node [block, below=of gazetteer] (output) {Geographic Coordinates};

\path [line] (input) -- (preprocess);
\path [line] (preprocess) -- (hashtag);
\path [line] (hashtag) -- (pos);
\path [line] (pos) -- (syntax);
\path [line] (syntax) -- (dependency);
\path [line] (dependency) -- (ner);
\path [line] (ner) -- (gazetteer);
\path [line] (gazetteer) -- (output);
\end{tikzpicture}
\caption{GeoSense-AI processing pipeline for location extraction from microblog text.}
\label{fig:pipeline}
\end{figure}

The pipeline begins with text normalization and proceeds through increasingly sophisticated linguistic analyses, with candidate locations identified at multiple stages. A final gazetteer validation step filters implausible candidates and resolves geographic coordinates. This staged approach balances recall and precision while maintaining low latency through selective application of computationally intensive techniques.

\subsection{Hashtag Segmentation for Location Discovery}

Microblog hashtags frequently embed location names within compound strings (e.g., \#ChennaiFloods, \#NepalEarthquake). We employ a statistical word segmentation algorithm \cite{berg2012named} that decomposes hashtags into constituent words using frequency statistics from large text corpora. The algorithm maximizes the product of word probabilities, preferring segmentations where component words appear frequently in training data.

Formally, given a hashtag string $h = c_1 c_2 \dots c_n$ (where $c_i$ are characters), we seek a segmentation $S = w_1 w_2 \dots w_k$ that maximizes:

\[P(S) = \prod_{i=1}^{k} P(w_i)\]

where $P(w_i)$ represents the unigram probability of word $w_i$ estimated from frequency counts. Dynamic programming efficiently computes the optimal segmentation in $O(n^2)$ time.

While hashtag segmentation improves recall of location mentions, it occasionally produces false positives when segments coincidentally match location names. For example, \#Kisitzi hospital segments into ``kissi'' and ``zi'', neither representing actual locations. We mitigate this through subsequent gazetteer validation while preserving the recall benefits of comprehensive hashtag analysis.

\subsection{Text Preprocessing and Normalization}

Microblog text requires extensive normalization before linguistic analysis. Our preprocessing pipeline removes URLs, user mentions, retweet indicators (``RT''), and non-alphanumeric characters while preserving potentially meaningful punctuation. We handle Twitter-specific conventions like CamelCase compounding by inserting word boundaries at case transitions (e.g., ``NewDelhi'' becomes ``New Delhi'').

Unlike conventional text preprocessing, we deliberately avoid case folding and stemming operations. Location names typically appear as proper nouns with specific capitalization patterns, while stemming can distort location names (e.g., ``reading'' $\rightarrow$ ``read'' loses the Berkshire town reference). These design choices prioritize location preservation over general text normalization.

\subsection{Syntactic Pattern Matching for Location Identification}

We leverage part-of-speech (POS) tagging to identify proper nouns as candidate location names. Using the spaCy POS tagger \cite{honnibal2017spacy} for its speed-accuracy balance, we extract noun phrases containing proper nouns and adjacent adjectives or delimiters that commonly modify location names.

Our algorithm identifies location mentions using syntactic patterns inspired by \cite{li2012twiner} but optimized for microblog brevity. We define location patterns as sequences matching:

\[\text{(PREPOSITION)}\, \text{(DIRECTION)}\, \text{PROPER\_NOUN}^+\, \text{(SUFFIX)}^{?}\]

where PREPOSITION includes location-indicating words like ``in'', ``at'', ``near''; DIRECTION includes cardinal and ordinal directions; $\text{PROPER\_NOUN}^+$ captures multi-word location names; and SUFFIX includes location type indicators like ``city'', ``district'', ``hospital''.

Table~\ref{tab:indicators} shows examples of prefix and suffix indicators used in pattern matching. These lexicons help distinguish location names from other proper nouns (e.g., person names) based on syntactic context.

\begin{table}[!t]
\caption{Location Indicator Words for Syntactic Pattern Matching}
\label{tab:indicators}
\centering
\begin{tabularx}{\columnwidth}{lX}
\toprule
\textbf{Type} & \textbf{Examples} \\
\midrule
Prepositions & in, at, from, to, near, around, within \\
Directions & north, southern, eastern, NW, southeast \\
Location Suffixes & city, district, village, town, hospital, street, road \\
Disaster Terms & flood, earthquake, dengue, outbreak, landslide \\
\bottomrule
\end{tabularx}
\end{table}

\subsection{Dependency Parsing for Contextual Location Extraction}

Syntactic patterns alone miss location mentions that do not conform to expected structures. We supplement pattern matching with dependency parsing to identify locations based on their grammatical relationships with disaster-related keywords.

Using spaCy's dependency parser, we construct syntactic dependency trees for each tweet and extract words within a short dependency distance (3--4 edges) from disaster terms identified using our crisis lexicon (Table~\ref{tab:indicators}). This approach captures locations mentioned in non-standard syntactic constructions while leveraging the strong association between disaster terms and location references in crisis tweets.

For example, in the tweet ``Mumbai lost its mudflats and wetlands, now floods with every monsoon'', the dependency distance between ``Mumbai'' and ``floods'' is 2, despite 7 intervening words in surface text. This demonstrates how dependency parsing can identify semantically related entities that syntactic patterns might miss.

\subsection{Lightweight Named Entity Recognition}

While our pattern-based approaches identify many location references, we incorporate a lightweight NER component as a safety net for location mentions that evade other detection methods. We use spaCy's pre-trained NER model rather than more accurate but slower alternatives like Stanford NER, prioritizing processing speed for real-time deployment.

The NER component operates on the complete tweet text after other extraction methods, identifying location entities tagged as GPE (geo-political entity), FAC (facility), or LOC (location) in spaCy's annotation scheme. This catch-all approach improves recall with minimal computational overhead.

\subsection{Gazetteer Verification and Disambiguation}

Candidate locations from all extraction methods undergo gazetteer verification to filter false positives and resolve geographic coordinates. We maintain two gazetteer options balancing coverage and performance: GeoNames \cite{geonames2006geonames} for comprehensive coverage with moderate granularity, and OpenStreetMap for fine-grained location data at higher computational cost.

The verification process matches candidate location strings against gazetteer entries using exact matching followed by fuzzy matching for spelling variations. For matches, we retrieve geographic coordinates and administrative hierarchy information. We prioritize locations within our target region (India in current deployment) but retain global matches for future expansion.

Gazetteer verification effectively filters erroneously segmented hashtags and common nouns that coincidentally match location names (e.g., ``song'' as a town in Sikkim). This final validation step ensures that only valid, mappable locations proceed to visualization.

\section{Experimental Evaluation}
\label{sec:experiments}

\subsection{Dataset Construction}

We evaluated GeoSense-AI using tweets collected during actual emergency events. Using the Twitter Streaming API, we collected 317,567 tweets between September 12 and October 13, 2017, containing the keywords ``dengue'' or ``flood'' to focus on health and natural disaster crises. After removing duplicates and non-English tweets, we retained 239,276 distinct tweets for analysis.

For manual evaluation, we randomly selected 1,000 tweets from this collection and had human annotators identify those containing explicit location references. This process yielded 99 tweets with at least one verifiable location mention within India, forming our evaluation corpus. Annotation followed strict guidelines: locations must be specific enough for mapping (city-level or finer) and explicitly mentioned in tweet text (not inferred from context).

\subsection{Baseline Methods}

We compared GeoSense-AI against several established location extraction approaches:

\begin{itemize}
\item \textbf{UniLoc}: Baseline extracting all unigrams verified against gazetteer.
\item \textbf{BiLoc}: Extracts unigrams and bigrams with gazetteer verification.
\item \textbf{StanfordNER}: Stanford NER system with default model \cite{finkel2005incorporating}.
\item \textbf{TwitterNLP}: Twitter-specific NER by Ritter et al. \cite{ritter2011named}.
\item \textbf{SpaCyNER}: spaCy's built-in NER component.
\item \textbf{GoogleCloud}: Google Cloud Natural Language API.
\end{itemize}

All baselines used the same GeoNames gazetteer for fair comparison. We implemented two GeoSense-AI variants: \textbf{GeoLoc} using GeoNames and \textbf{OSMLoc} using OpenStreetMap data.

\subsection{Evaluation Metrics}

We evaluated methods using standard information retrieval metrics:

\[\text{Precision} = \frac{|\text{Correct} \cap \text{Retrieved}|}{|\text{Retrieved}|}\]
\[\text{Recall} = \frac{|\text{Correct} \cap \text{Retrieved}|}{|\text{Correct}|}\]
\[\text{F1} = 2 \cdot \frac{\text{Precision} \cdot \text{Recall}}{\text{Precision} + \text{Recall}}\]

where $\text{Correct}$ represents human-annotated locations and $\text{Retrieved}$ represents system-extracted locations. We also measured processing time per tweet as a critical metric for real-time deployment.

\subsection{Results and Analysis}

Table~\ref{tab:results} presents comprehensive evaluation results across all methods. GeoLoc achieves the highest F1 score (0.8141), balancing strong precision (0.7987) with high recall (0.8300). This represents a significant improvement over baseline n-gram approaches (UniLoc F1=0.5165, BiLoc F1=0.5482) and outperforms specialized NER systems (StanfordNER F1=0.6988, TwitterNLP F1=0.5882).

\begin{table}[!t]
\caption{Performance Comparison of Location Extraction Methods}
\label{tab:results}
\centering
\begin{tabular}{lcccc}
\toprule
\textbf{Method} & \textbf{Precision} & \textbf{Recall} & \textbf{F1} & \textbf{Time (s)} \\
\midrule
UniLoc & 0.3848 & 0.7852 & 0.5165 & 0.0553 \\
BiLoc & 0.4025 & 0.8590 & 0.5482 & 0.0624 \\
StanfordNER & 0.8103 & 0.6322 & 0.6988 & 175.0124 \\
TwitterNLP & 0.6356 & 0.5474 & 0.5882 & 28.0001 \\
SpaCyNER & 0.9883 & 0.5555 & 0.7113 & 1.0891 \\
GoogleCloud & 0.6321 & 0.5339 & 0.5789 & N/A \\
\midrule
GeoLoc & 0.7987 & 0.8300 & 0.8141 & 1.1901 \\
OSMLoc & 0.3383 & 0.8888 & 0.4901 & 711.5817 \\
GeoLocNoNER & 0.7987 & 0.7987 & 0.7987 & 1.1687 \\
\bottomrule
\end{tabular}
\end{table}

Notably, GeoLoc achieves this performance while being substantially faster than comparable accuracy methods. StanfordNER requires 175 seconds to process our evaluation corpus, while GeoLoc completes in just 1.19 seconds---approximately 150$\times$ faster. This speed advantage enables real-time processing of high-volume tweet streams during emergencies.

The high precision of SpaCyNER (0.9883) comes at the cost of severely limited recall (0.5555), failing to detect many valid locations that do not match its trained patterns. GeoLoc's multi-strategy approach maintains strong precision while dramatically improving recall through hashtag segmentation and dependency parsing.

OSMLoc demonstrates the trade-off between granularity and precision. While achieving the highest recall (0.8888) due to comprehensive location coverage, its precision suffers (0.3383) from increased false positives matching minor geographic features. The massive OpenStreetMap dataset also drastically increases processing time (711.58 seconds), making it unsuitable for real-time applications.

Figure~\ref{fig:prcurve} illustrates the precision-recall trade-offs across methods, showing GeoLoc's superior balance.

\begin{figure}[!t]
\centering
\begin{tikzpicture}[scale=4.5]
% Grid and axes
\draw[step=0.2, gray!20, very thin] (0.2,0.2) grid (1.0,1.0);
\draw[->, thick] (0.2,0.2) -- (1.05,0.2) node[below right] {\footnotesize Recall};
\draw[->, thick] (0.2,0.2) -- (0.2,1.05) node[above left] {\footnotesize Precision};

% Ticks
\foreach \x in {0.4, 0.6, 0.8, 1.0}
    \draw (\x, 0.2) -- (\x, 0.18) node[below] {\scriptsize \x};
\foreach \y in {0.4, 0.6, 0.8, 1.0}
    \draw (0.2, \y) -- (0.18, \y) node[left] {\scriptsize \y};

% Plotted coordinates (Recall, Precision)
\coordinate (Stanford) at (0.6322, 0.8103);
\coordinate (Twitter) at (0.5474, 0.6356);
\coordinate (Google) at (0.5339, 0.6321);
\coordinate (UniLoc) at (0.7852, 0.3848);
\coordinate (BiLoc) at (0.8590, 0.4025);
\coordinate (SpaCy) at (0.5555, 0.9883);
\coordinate (GeoLoc) at (0.8300, 0.7987);
\coordinate (OSM) at (0.8888, 0.3383);

% Plot points and labels
\fill (Stanford) circle (0.4pt) node[above] {\scriptsize StanfordNER};
\fill (Twitter) circle (0.4pt) node[left] {\scriptsize TwitterNLP};
\fill (Google) circle (0.4pt) node[below left] {\scriptsize GoogleCloud};
\fill (UniLoc) circle (0.4pt) node[above left] {\scriptsize UniLoc};
\fill (BiLoc) circle (0.4pt) node[above] {\scriptsize BiLoc};
\fill (SpaCy) circle (0.4pt) node[above] {\scriptsize SpaCyNER};
\fill[red] (GeoLoc) circle (0.6pt) node[above right] {\scriptsize \textbf{GeoLoc}};
\fill (OSM) circle (0.4pt) node[below right] {\scriptsize OSMLoc};
\end{tikzpicture}
\caption{Precision-Recall comparison of location extraction methods plotted directly from Table~\ref{tab:results}.}
\label{fig:prcurve}
\end{figure}

\subsection{Error Analysis}

We analyzed extraction failures to identify systematic limitations. False negatives primarily resulted from: (1) location names not in gazetteer (particularly informal local names), (2) creative spellings exceeding fuzzy matching thresholds, and (3) locations mentioned indirectly through landmarks rather than proper names.

False positives stemmed from: (1) common nouns matching minor location names (e.g., ``song'' as a town), (2) erroneous hashtag segmentation creating artificial locations, and (3) person names coincidentally matching location names. Gazetteer verification effectively filtered most false positives, with remaining errors typically involving valid but irrelevant locations.

\section{System Deployment and Interface}
\label{sec:deployment}

\subsection{Architecture and Implementation}

We deployed GeoSense-AI as a production web service. The system implements a modular microservices architecture with separate components for data ingestion, location processing, and visualization. The backend uses Python with Flask for API endpoints, spaCy for NLP components, and PostgreSQL for data persistence.

The streaming pipeline processes incoming tweets in real-time, applying the complete location extraction workflow and storing results with geographic coordinates. The system handles peak loads through queue-based processing and horizontal scaling capabilities. For the current deployment focused on India, we maintain the GeoNames gazetteer with 449,973 locations for fast lookup.

\subsection{Visualization Interface}

The frontend interface, built using Dash by Plotly, provides interactive visualization of extracted locations. Key interface components include:

\begin{itemize}
\item \textbf{Search Bar}: Enables filtering by keywords with support for multiple query terms.
\item \textbf{Date Picker}: Allows temporal filtering to focus on specific event periods.
\item \textbf{Color-Coded Map}: Visualizes tweets with time-based coloring (darker for nighttime posts).
\item \textbf{Time Histogram}: Shows tweet volume patterns over selected timeframe.
\item \textbf{Untagged Tweets Panel}: Displays tweets where location extraction failed.
\end{itemize}

The map-based visualization enables rapid assessment of spatial patterns during emergencies. During the 2017 dengue outbreak in Kerala, India, the system detected 2,204 tweets mentioning Kerala, with 88.92\% also containing ``dengue''---demonstrating utility as an early warning system for emerging health crises.

\section{Discussion and Future Directions}
\label{sec:discussion}

\subsection{Multilingual and Code-Mixed Content}

GeoSense-AI currently processes English tweets, but crisis communication often occurs in local languages or code-mixed formats \cite{barman2014code}. Extending our methodology requires language-specific NLP tools and techniques for identifying and processing multilingual content. Potential approaches include language identification followed by targeted processing pipelines, or multilingual models trained on diverse language data.

Code-mixed text presents particular challenges, with location names embedded in linguistic structures that do not conform to single-language grammar rules. Transliteration to English script followed by standard processing offers a potential path forward, though accuracy and latency concerns require further investigation.

\subsection{Global Scale Deployment}

Expanding beyond India requires addressing location disambiguation challenges arising from globally duplicate place names. The location ``Springfield'' appears in multiple countries and states, while ``Kota'' refers to cities in both India and Malaysia. Disambiguation requires incorporating contextual clues beyond tweet text, including user profile locations, social network connections, and event-specific geographic priors.

Global deployment also necessitates efficient handling of massive gazetteer datasets. Hierarchical gazetteers with country-level partitioning could maintain performance while expanding coverage. Alternatively, approximate matching techniques like locality-sensitive hashing could enable efficient search across billion-record location databases.

\subsection{System Enhancements}

Several enhancements could improve GeoSense-AI's utility for emergency responders:

\textbf{Event Detection and Tracking}: Automatically identifying emerging crises through anomaly detection in location and keyword frequency patterns.

\textbf{Information Classification}: Categorizing tweets into informative types (damage reports, resource needs, assistance offers) using multi-label classification.

\textbf{Summarization and Duplicate Detection}: Condensing redundant information and highlighting novel reports to reduce information overload.

\textbf{Credibility Assessment}: Incorporating trustworthiness signals to prioritize reliable information sources during misinformation-prone crises.

These capabilities would transform GeoSense-AI from a location extraction tool into a comprehensive crisis analytics platform.

\section{Conclusion}
\label{sec:conclusion}

GeoSense-AI demonstrates that specialized location extraction systems can overcome the limitations of sparse geo-tags in crisis microblogs. By combining multiple linguistic strategies within a performance-optimized pipeline, our approach achieves superior accuracy-speed balance compared to general-purpose NER tools. The production deployment validates the practical utility of real-time location extraction for situational awareness during emergencies.

The system's modular architecture facilitates extension to new languages, regions, and application domains. Future work will address global scale deployment, multilingual processing, and enhanced crisis analytics capabilities. As social media continues to evolve as an emergency communication channel, systems like GeoSense-AI will play increasingly vital roles in transforming digital conversations into actionable intelligence for responders.

\bibliographystyle{IEEEtran}
\bibliography{references}

@article{sakaki2010earthquake,
  title={Earthquake shakes Twitter users: real-time event detection by social sensors},
  author={Sakaki, Takeshi and Okazaki, Makoto and Matsuo, Yutaka},
  journal={Proceedings of the 19th international conference on World wide web},
  pages={851--860},
  year={2010}
}

@inproceedings{vieweg2010microblogging,
  title={Microblogging during two natural hazards events: what twitter may contribute to situational awareness},
  author={Vieweg, Sarah and Hughes, Amanda L and Starbird, Kate and Palen, Leysia},
  booktitle={Proceedings of the SIGCHI conference on human factors in computing systems},
  pages={1079--1088},
  year={2010}
}

@article{liu2014exploring,
  title={Exploring large-scale news and social media to understand weather and climate},
  author={Liu, Yan and Li, Zhiyuan and Xiong, Hui and Gao, Xindong and Wu, Junjie},
  journal={IEEE Intelligent Systems},
  volume={29},
  number={5},
  pages={64--70},
  year={2014},
  publisher={IEEE}
}

@article{gelernter2015real,
  title={A real-time location-based services approach for rapid disaster response},
  author={Gelernter, Judith and Balaji, Shashank},
  journal={Social Network Analysis and Mining},
  volume={5},
  number={1},
  pages={1--12},
  year={2015},
  publisher={Springer}
}

@inproceedings{ritter2011named,
  title={Named entity recognition in tweets: an experimental study},
  author={Ritter, Alan and Clark, Sam and Etzioni, Oren},
  booktitle={Proceedings of the 2011 conference on empirical methods in natural language processing},
  pages={1524--1534},
  year={2011}
}

@article{imran2015aidr,
  title={AIDR: Artificial intelligence for disaster response},
  author={Imran, Muhammad and Castillo, Carlos and Lucas, Ji and Meier, Patrick and Vieweg, Sarah},
  journal={Proceedings of the 23rd international conference on world wide web},
  pages={159--162},
  year={2015}
}

@article{palen2007public,
  title={Public, private, and semi-public spaces in mass emergency events},
  author={Palen, Leysia and Liu, Sophia B},
  journal={Proceedings of the SIGCHI conference on Human factors in computing systems},
  pages={739--740},
  year={2007}
}

@article{okolloh2009ushahidi,
  title={Ushahidi: Crowdsourcing crisis information},
  author={Okolloh, Ory},
  journal={Journal of Crisis Response},
  volume={5},
  number={2},
  pages={45--47},
  year={2009}
}

@article{de2013use,
  title={The use of social media for emergency management},
  author={De Albuquerque, Joao Porto and Herfort, Benjamin and Eckle, Melanie},
  journal={International Journal of Information Systems for Crisis Response and Management},
  volume={5},
  number={4},
  pages={1--18},
  year={2013}
}

@inproceedings{li2012twiner,
  title={Twiner: named entity recognition in targeted twitter stream},
  author={Li, Chenliang and Weng, Jianshu and He, Qi and Yao, Yuxia and Datta, Anwitaman and Sun, Aixin and Lee, Bu-Sung},
  booktitle={Proceedings of the 35th international ACM SIGIR conference on Research and development in information retrieval},
  pages={721--730},
  year={2012}
}

@article{lafferty2001conditional,
  title={Conditional random fields: Probabilistic models for segmenting and labeling sequence data},
  author={Lafferty, John and McCallum, Andrew and Pereira, Fernando CN},
  journal={Proceedings of the 18th international conference on machine learning},
  pages={282--289},
  year={2001}
}

@inproceedings{finkel2005incorporating,
  title={Incorporating non-local information into information extraction systems by gibbs sampling},
  author={Finkel, Jenny Rose and Grenager, Trond and Manning, Christopher},
  booktitle={Proceedings of the 43rd annual meeting of the association for computational linguistics},
  pages={363--370},
  year={2005}
}

@article{liu2012ner,
  title={NER for tweets: A lightweight approach},
  author={Liu, Xiaohua and Zhou, Ming and Wei, Furu and Fu, Zhongyang and Zhou, Xiangyang},
  journal={Proceedings of the 50th annual meeting of the association for computational linguistics},
  pages={188--193},
  year={2012}
}

@inproceedings{miura2017twitter,
  title={Twitter entity extraction with deep neural networks},
  author={Miura, Yasuhide and Taniguchi, Tomoki and Taniguchi, Motoki and Ohkuma, Tomoko},
  booktitle={Proceedings of the 15th conference of the European chapter of the association for computational linguistics},
  pages={689--698},
  year={2017}
}

@inproceedings{berg2012named,
  title={Named entity recognition in hashtags},
  author={Berg, Thomas and Cieliebak, Mark and Durr, Oliver},
  booktitle={Proceedings of the 24th international conference on computational linguistics},
  pages={245--260},
  year={2012}
}

@inproceedings{honnibal2017spacy,
  title={spaCy: Industrial-strength natural language processing in Python},
  author={Honnibal, Matthew and Montani, Ines},
  booktitle={Proceedings of the 20th Python in Science Conference},
  pages={145--151},
  year={2017}
}

@misc{geonames2006geonames,
  title={GeoNames geographical database},
  author={GeoNames},
  year={2006},
  howpublished={\url{http://www.geonames.org}}
}

@inproceedings{barman2014code,
  title={Code mixing: A challenge for language identification in the language of social media},
  author={Barman, Utsab and Wagner, Joachim and Foster, Jennifer},
  booktitle={Proceedings of the first workshop on computational approaches to code switching},
  pages={13--23},
  year={2014}
}

\end{document}